\documentclass[10pt,twocolumn,letterpaper]{article}

\usepackage[pagenumbers]{cvpr} 

\usepackage[dvipsnames]{xcolor}

\usepackage{tikz}
\newcommand{\R}{\mathbb{R}}
\usetikzlibrary{positioning,shapes,arrows}

\usepackage{booktabs}
\usepackage{makecell}
\usepackage{subcaption}
\definecolor{cvprblue}{rgb}{0.21,0.49,0.74}
\usepackage[pagebackref,breaklinks,colorlinks,citecolor=cvprblue]{hyperref}

\def\paperID{473} 
\def\confName{CVPR}
\def\confYear{2024}

\newcommand{\lora} {LoRA}
\newcommand{\name} {ZipLoRA}

\title{\name{}: Any Subject in Any Style by Effectively Merging LoRAs}

\author{Viraj Shah$^{1,2}$ \quad Nataniel Ruiz$^{1}$ \quad Forrester Cole$^{1}$ \quad Erika Lu$^{1}$ \\ Svetlana Lazebnik$^{2}$ \quad Yuanzhen Li$^{1}$ \quad Varun Jampani$^{1}$ \\
${^1}~$Google Research \quad $^{2}~$UIUC
}

\begin{document}
\twocolumn[{%
	\renewcommand\twocolumn[1][]{#1}%
	\maketitle
	\vspace{-2.8em}
	\begin{center}
		\includegraphics[width=0.92\linewidth]{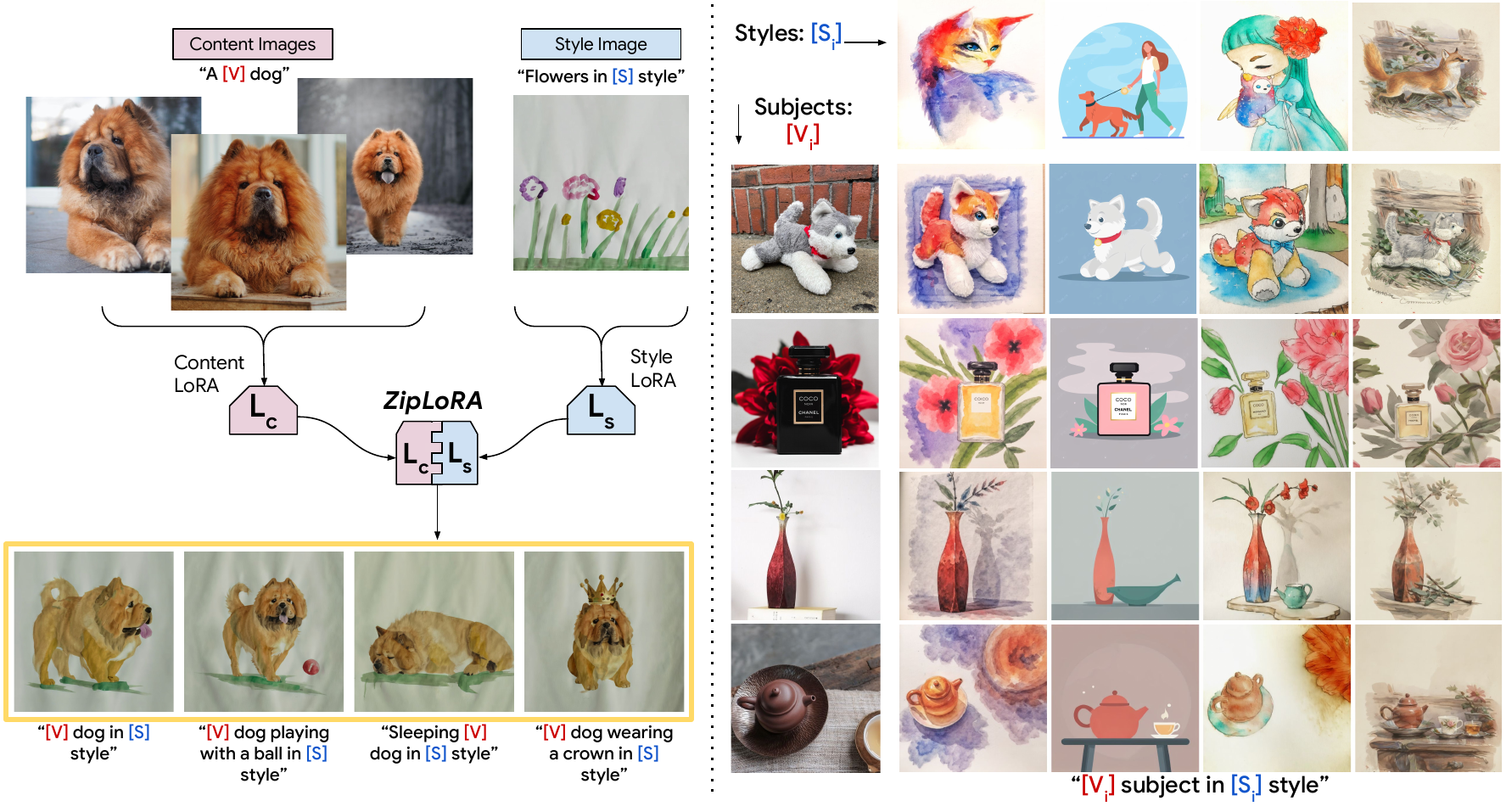}
	\end{center}
	\vspace{-1.0em}
	\captionof{figure}{By effectively merging independently trained style and content LoRAs, our proposed method \textbf{\name{}} is able to generate \textit{any user-provided subject in any user-provided style}, providing unprecedented control over personalized creations using diffusion models.
		}
	\label{fig:teaser}
	\vspace{1em}
}]
\maketitle

\begin{abstract}
\vspace{-6mm}
Methods for finetuning generative models for concept-driven personalization generally achieve strong results for subject-driven or style-driven generation. Recently, low-rank adaptations (\lora{}) have been proposed as a parameter-efficient way of achieving concept-driven personalization. While recent work explores the combination of separate LoRAs to achieve joint generation of learned styles and subjects, existing techniques do not reliably address the problem, so that either subject fidelity or style fidelity are compromised. We propose \textbf{\name{}}, a method to cheaply and effectively merge independently trained style and subject LoRAs in order to achieve generation of \textbf{any user-provided subject in any user-provided style}. Experiments on a wide range of subject and style combinations show that \name{} can generate compelling results with meaningful improvements over baselines in subject and style fidelity while preserving the ability to recontextualize.
\end{abstract}

\section{Introduction}
\label{sec:intro}
Recently, diffusion models~\cite{Ho2020DenoisingDP, Song2020DenoisingDI, Rombach2021HighResolutionIS} have allowed for impressive image generation quality with their excellent understanding of diverse artistic concepts and enhanced controllability due to multi-modal conditioning support (with text being the most popular mode). 
The usability and flexibility of generative models has further progressed with a wide variety of personalization approaches, such as DreamBooth~\cite{ruiz2023dreambooth} and StyleDrop~\cite{sohn2023styledrop}. These approaches fine-tune a base diffusion model on the images of a specific concept to produce novel renditions in various contexts. Such concepts can be a specific object, person, or artistic style.

While personalization methods have been used for subjects and styles independently, a key unsolved problem is to generate a specific user-provided \emph{subject} in a specific user-provided \emph{style}. For example, an artist may wish to render a specific person in their personal style, learned through examples of their own work. A user may wish to generate images of their child's favorite plush toy, in the style of the child's watercolor paintings. Moreover, if this is achieved two problems are simultaneously solved: (1) the task of representing any given subject in any style, and (2) the problem of controlling diffusion models through images rather than text, which can be imprecise and unsuitable for certain generation tasks. Finally, we can imagine a large-scale application of such a tool, where a bank of independently learned styles and subjects are shared and stored online. The task of arbitrarily rendering \textit{any subject in any style} is an open research problem that we seek to address.

A pitfall of recent personalization methods is that many finetune all of the parameters of a large base model, which can be costly. Parameter Efficient Fine-Tuning (PEFT) approaches allow for fine-tuning models for concept-driven personalization with much lower memory and storage budgets. Among the various PEFT approaches, Low Rank Adaptation (LoRA)~\cite{hu2022lora} has emerged as a favored method for researchers and practitioners alike due to its versatility. 
LoRA learns low-rank factorized weight matrices for the attention layers (these learned weights are themselves commonly referred to as ``LoRAs'').
By combining LoRA and algorithms such as DreamBooth~\cite{ruiz2023dreambooth}, the learned subject-specific LoRA weights enable the model to generate the subject with semantic variations.
    
With the growing popularity of LoRA personalization, there have been attempts to merge LoRA weights, specifically by performing a linear combination of subject and style LoRAs, with variable coefficients~\cite{simo_lora}. This allows for a control over the ``strength'' of each LoRA, and users sometimes are able, through careful grid search and subjective human evaluation, to find a combination that allows for accurate portrayal of the subject under the specific style. This method lacks robustness across style and subject combinations, and is also incredibly time consuming.

In this work, we propose \textit{\name{}}, a simple yet effective method to generate any subject in any style by cheaply merging independently trained LoRAs for subject and style. Note that since we aim to achieve custom stylization of a given subject, we focus specifically on merging two LoRAs (one for subject and one for style).
Our approach works consistently on a wide variety of subject and style LoRAs without enforcing any restriction on the way these are trained. This allows users and artists to easily combine publicly available subject and style LoRAs of their choice. \name{} is hyperparameter-free, i.e. it does not require manual tuning of any hyperparameters or merger weights.

Our approach is based on two important observations. \textbf{(1)} LoRA weights for different layers $\Delta W_i$ (where $i$ denotes the layer) are sparse. \textit{i.e.}, most of the elements in $\Delta W_i$ have very small magnitude, and have little effect on generation quality and fidelity. \textbf{(2)} Columns of the weight matrices of two independently trained LoRAs may have varying levels of ``alignment'' between each other, as measured by cosine similarity, for example. We find that directly summing columns that are highly aligned degrades performance of the merged model. 

Based on these observations, we hypothesize that a method that operates akin to a zipper, aiming to reduce the quantity of similar-direction sums while preserving the content and style generation properties of the original LoRAs will yield more robust, higher-quality merges. Much like a zipper seamlessly joins two sides of a fabric, our proposed optimization-based approach finds a disjoint set of merger coefficients for blending the subject and style LoRAs, ensuring that the merge adeptly captures both subject and style. Our optimization process is lightweight and significantly improves the merging performance on challenging content-style combinations, where the two LoRAs are highly aligned.

While our approach is independent of the model architecture, we further observe that the recently released Stable Diffusion XL (SDXL) model~\cite{Podell2023SDXLIL} exhibits strong style learning properties, comparable to results shown by StyleDrop~\cite{sohn2023styledrop} on Muse~\cite{chang2023muse}. Specifically, unlike previous versions of Stable Diffusion~\cite{Rombach2021HighResolutionIS}, SDXL is able to learn styles using just a single exemplar image by following a DreamBooth protocol~\cite{ruiz2023dreambooth} without any human feedback. This property makes our method particularly effective when applied to SDXL.
We summarize our contributions as follows:
\begin{itemize}
    \item We demonstrate some key observations about current text-to-image diffusion models and personalization methods, particularly in relation to style personalization. We further examine the sparsity of concept-personalized LoRA weight matrix coefficients and the prevalence and deleterious effect of highly aligned columns for LoRA matrices.
    \item Using these insights we propose \textbf{\name{}}, a simple optimization method that allows for effective merging of independently trained style and subject LoRAs to allow for the generation of \textit{any subject in any style.}
    \item We demonstrate the effectiveness of our approach on a variety of image stylization tasks, including content-style transfer and recontextualization. We also demonstrate that \name{} outperforms existing methods of merging LoRAs as well as other baseline approaches.
\end{itemize}
\section{Related Work}
\label{sec:related_work}
\begin{figure*}[tp]
	\centering
	\includegraphics[width=1\linewidth]{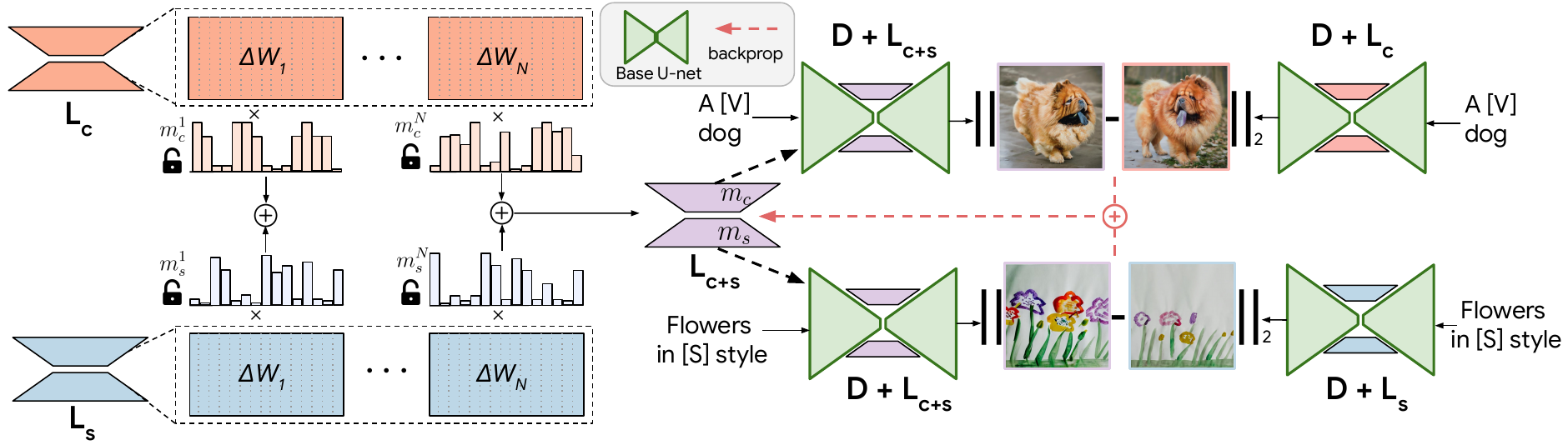}
	\caption{\textbf{Overview of \name{}}. Our method learns mixing coefficients for each column of $\Delta W_i$ for both style and subject LoRAs. It does so by \textbf{(1)} minimizing the difference between subject/style images generated by the mixed LoRA and original subject/style LoRA models, while \textbf{(2)} minimizing the cosine similarity between the columns of content and style LoRAs. In essence, the zipped LoRA tries to conserve the subject and style properties of each individual LoRA, while minimizing signal interference of both LoRAs.}
	\label{fig:method_overview}
\end{figure*}
\noindent\textbf{Image Stylization.} Image-based style transfer is an area of research dating back at least 20 years~\cite{efros2001image,Hertzmann2001ImageA}. Great advances in arbitrary style transfer was achieved by the convolutional neural network-based approaches~\cite{gatys2016image,johnson2016perceptual,huang2017arbitrary, WCT-NIPS-2017,park2019arbitrary}. 
Generative models such as GANs~\cite{Karras2019ASG, Karras2020AnalyzingAI, Karras2020TrainingGA} can also be used as a prior for image stylization tasks~\cite{Menon2020PULSESP, Bora2017CompressedSU,Wang2021TowardsRB}. Many recent GAN-based approaches achieve successful one-shot stylizations~\cite{Liu2021BlendGANIG, Gal2021StyleGANNADACD, oneshotadaption, Ojha2021FewshotIG, Wang2022CtlGANFA, Kwon2022OneShotAO, zhu2022mind, jojogan, multistylegan, pastiche} by fine-tuning a pre-trained GAN for a given reference style. However, these methods are limited to images from only a single domain (such as faces). Further, most existing GANs do not provide any direct, text-based control over the semantics of the output, thus they cannot produce the reference subject in novel contexts. Methods such as~\cite{Kwon2021CLIPstylerIS, Jandial2023GathaRL,Gal2021StyleGANNADA} attempt to modulate the style of the content image using the text description, however, they do not support a style reference image like our approach, and do not provide re-contextualization capability. Compared to older generative models, diffusion models~\cite{Ho2020DenoisingDP, Song2020DenoisingDI, Rombach2021HighResolutionIS} offer superior generation quality and text-based control; however, to date, it has been difficult to use them for one-shot stylization driven by image examples. Ours is one of the first works demonstrating the use of diffusion models for high-quality example-based stylization combined with an ability to re-contextualize to diverse scenarios.

\noindent\textbf{Fine-tuning of Diffusion Models for Custom Generation.} In the evolving field of text-to-image (T2I) model personalization, recent studies have introduced various methods to fine-tune large-scale T2I diffusion models for depicting specific subjects based on textual descriptions. Techniques like Textual Inversion~\cite{gal2022textual} focus on learning text embeddings, while DreamBooth~\cite{ruiz2023dreambooth} fine-tunes the entire T2I model for better subject representation. Later methods aim to optimize specific parts of the networks~\cite{han2023svdiff,kumari2022customdiffusion}. Additionally, techniques like LoRA~\cite{hu2022lora} and StyleDrop~\cite{sohn2023styledrop} concentrate on optimizing low-rank approximations and a small subset of weights, respectively, for style personalization. DreamArtist~\cite{dong2023dreamartist} introduces a novel one-shot personalization method using a positive-negative prompt tuning strategy. While these fine-tuning approaches yield high-quality results, they typically are limited to learning only one concept (either subject or style). One exception is Custom Diffusion~\cite{kumari2022customdiffusion}, which attempts to learn multiple concepts simultaneously. However, Custom Diffusion requires expensive joint training from scratch and still yields inferior results when used for stylization as it fails to disentangle the style from the subject.

\noindent\textbf{Combining LoRAs.} Combining different LoRAs remain under-explored in the literature particularly from the point of view of fusing style and the subject concepts. Ryu \cite{simo_lora} shows a method to combine independently trained LoRAs by weighed arithmetic summation. In~\cite{gu2023mixofshow}, authors discuss fusing multiple concept LoRAs using gradient fusion strategy, however, it is an expensive method that requires retraining the entire model. Further, since it uses a custom LoRA variant referred to as ED-LoRA, it lacks the flexibility to combine freely available pre-trained LoRAs. It also relies on regional prompting that uses different prompts for different regions of the image -- a trick that is unsuitable for subject-style merge since the style cannot be localized to any one location in the image. A concurrent work discusses a strategy to obtain Mixture of Experts by combining multiple LoRAs using a gating function~\cite{wu2024mole}. However, it focuses only on the ability to generate the individual concepts separately, and does not consider the problem of combined generation, \textit{i.e.} generating multiple different concepts (such as object and style) together in a single image.

\section{Methods}
\label{sec:method}
\subsection{Background}
\noindent \textbf{Diffusion Models}~\cite{Ho2020DenoisingDP, Song2020DenoisingDI, Rombach2021HighResolutionIS} are state-of-the-art generative models known for their high-quality, photorealistic image synthesis. Their training comprises two phases: a forward process, where an image transitions into a Gaussian noise through incremental Gaussian noise addition, and a reverse process, reconstructing the original data from the noise. The reverse process is typically learnt using an U-net with text conditioning support enabling text-to-image generation at the time of inference. In our work, we focus on widely used latent diffusion model~\cite{Rombach2021HighResolutionIS} which learns the diffusion process in the latent space instead of image space. In particular, we use Stable Diffusion XL v1~\cite{Podell2023SDXLIL} for all our experiments.

\noindent \textbf{LoRA Fine-tuning. }LoRA (Low-Rank Adaptation) is a method for efficient adaptation of Large Language and Vision Models to a new downstream task~\cite{hu2022lora,simo_lora}. The key concept of LoRA is that the weight updates $\Delta W$ to the base model weights $W_0 \in \R^{m \times n}$ during fine-tuning have a ``low intrinsic rank," thus the update $\Delta W$ can be decomposed into two low-rank matrices $B \in \R^{m \times r}$ and $A\in \R^{r \times n}$ for efficient parameterization with $\Delta W = BA$. Here, $r$ represents the intrinsic rank of $\Delta W$ with $r << min(m,n)$. During training, only $A$ and $B$ are updated to find suitable $\Delta W = BA$, while keeping $W_0$ constant. For inference, the updated weight matrix $W$ can be obtained as $W = W_0 + BA$. Due to its efficiency, LoRA is widely used for fine-tuning open-sourced diffusion models.
\subsection{Problem Setup}
In this work, we aim to produce accurate renditions of a custom object in a given reference style by merging 
LoRA weights obtained by separately fine-tuning a given text-to-image diffusion model on a few reference images of the object/style.

We start with a base diffusion model represented as $D$ with pre-trained weights $W^{(i)}_0$ with $i$ as layer index. One can adapt the base model $D$ to any given concept by simply adding the corresponding set of LoRA weights $L_x \{\Delta W_x^{(i)}\}$ to the model weights. We represent it as: $D_{L_x} = D \oplus L_x = W_0 + \Delta W_x$. We drop the superscript $(i)$ for simplicity since our operations are applied over all the LoRA-enabled weight matrices of our base model $D$.

We are given two independently trained set of LoRAs $L_c= \{\Delta W_c^{(i)}\}$ and $L_s= \{\Delta W_s^{(i)}\}$ for our base model $D$, and we aim to find a merged LoRA $L_m = \{\Delta W^{(i)}_m\} = \mathrm{Merge}(L_c, L_s)$ that can combine the effects of both the individual LoRAs in order to stylize the given object in a desired reference style. 

\noindent\textbf{Direct Merge. }LoRA is popularly used as a plug-and-play module on top of the base model, thus a most common way to combine multiple LoRAs is a simple linear combination~\cite{simo_lora}:
\begin{align}
L_m = L_c + L_s \implies \Delta W_m =  w_c\cdot \Delta W_c + w_s\cdot \Delta W_s, 
\label{eq:direct}
\end{align}
where $w_c$ and $w_s$ are coefficients of content and style LoRAs, respectively, which allow for a control over the ``strength'' of each LoRA. For a given subject and style LoRA, one may be able to find a particular combination of $w_c$ and $w_s$ that allows for accurate stylization through careful grid search and subjective human evaluation, but this method is not robust and very time consuming. To this end, we propose a hyperparameter-free approach that does not require this onerous process.

\subsection{\name{}}
\begin{figure*}
\centering
\begin{subfigure}{.42\textwidth}
  \centering
  \includegraphics[width=0.95\linewidth]{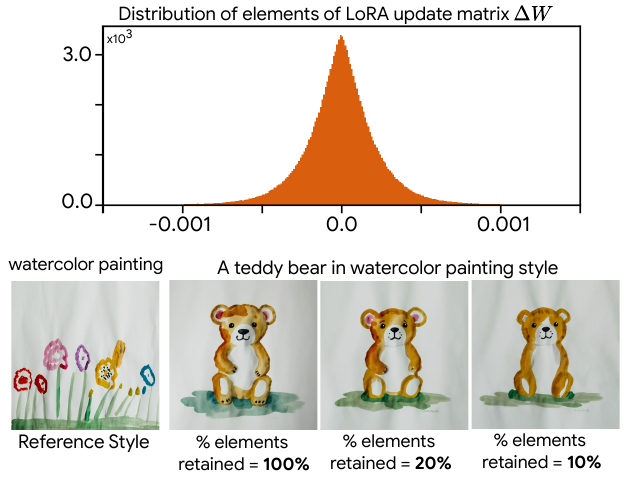}
    \caption{\scriptsize\textbf{LoRA weight matrices are sparse.} }
    \label{fig:sparse}
  
\end{subfigure}%
\quad
\begin{subfigure}{.54\textwidth}
  \centering
  \includegraphics[width=0.98\linewidth]{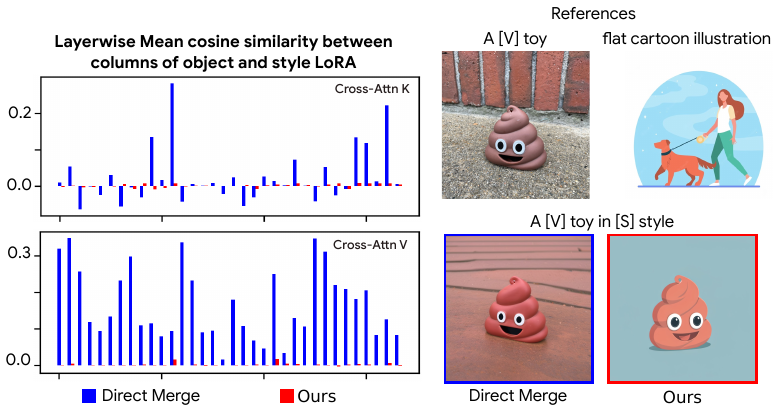}
  \vspace{2pt}
    \caption{\scriptsize\textbf{Highly aligned LoRA weights merge poorly. }}
    \label{fig:insight}
\end{subfigure}
\caption{Key insights of our approach: \textbf{(a)} Most of the elements in $\Delta W$ have a magnitude very close to zero, and can be conveniently thrown away without affecting the generation quality of the fine-tuned model. \textbf{(b)} When LoRA weight columns are highly aligned, a direct merge obtains subpar results. Instead, our approach minimizes the mean cosine similarity between the columns of the LoRA updates across the layers.}
\vspace{-1em}
\end{figure*}

Our approach builds on two interesting insights:

\noindent\textbf{(1) LoRA update matrices are sparse.} 
We observe that the update matrices $\Delta W$ for different LoRA layers are sparse, \textit{i.e.}, most of the elements in $\Delta W$ have a magnitude very close to zero, and thus have little impact on the output of the fine-tuned model. For each layer, we can sort all the elements by their magnitude and zero out the lowest up to a certain percentile. We depict the distribution of elements of $\Delta W_i^{m \times n}$ in Fig.~\ref{fig:sparse}, along with samples generated after zeroing out 80\% and 90\% of the lowest-magnitude elements of weight update matrix $\Delta W$ for all the layers. As can be seen, the model performance is unaffected even when 90\% of the elements are thrown away. This observation follows from the fact that the rank of $\Delta W$ is very small by design, thus the information contained in most columns of $\Delta W$ is redundant. 

\noindent\textbf{(2) Highly aligned LoRA weights merge poorly.} 
Columns of the weight matrices of two independently trained LoRAs may contain information that is not disentangled, \textit{i.e.}, the cosine similarity between them can be non-zero. We observe that the extent of alignment between the columns of LoRA weights plays a significant role in determining the quality of resulting merge: if we directly add the columns with non-zero cosine similarity to each other, it leads to superimposition of their information about the individual concepts, resulting in the loss of the ability of the merged model to synthesize input concepts accurately. We further observe that such loss of information is avoided when the columns are orthogonal to each other with cosine similarity equal to zero.

Note that each weight matrix represents a linear transformation defined by its columns, so it is intuitive that the merger would retain the information available in these columns only when the columns that are being added are orthogonal to each other. For most content-style LoRA pairs the cosine similarities are non-zero, resulting in signal interference when they are added directly. In Fig.~\ref{fig:insight} we show the mean cosine similarity values for each layer of the last U-net block for a particular content-style pair before and after applying ZipLoRA. One can see high non-zero cosine similarity values for the direct merge which results in poor stylization quality. On the other hand, ZipLoRA reduces the similarity values significantly to achieve a superior result.

To prevent signal interference during the merger, we multiply each column with a learnable coefficient such that the orthogonality between the columns can be achieved. The fact that LoRA updates are sparse allows us to neglect certain columns from each LoRA, thus facilitating the task of minimizing interference. As shown in Fig.~\ref{fig:method_overview}, we introduce a set of merger coefficient vectors $m_c$ and $m_s$ for each LoRA layer of the content and style LoRAs, respectively:
\begin{align}
  L_m = \mathrm{Merge}(L_c, L_s, m_c, m_s)  \nonumber \\  \implies \Delta W_m = m_c \otimes \Delta W_c  + m_s \otimes W_s,
  \label{eq:merge}
\end{align}
where $\otimes$ represents element-wise multiplication between $\Delta W$ and broadcasted merger coefficient vector $m$ such that $j^{th}$ column of $\Delta W$ gets multiplied with $j^{th}$ element of $m$. The dimensionalities of $m_c$ and $m_s$ are equal to the number of columns in corresponding $\Delta W$, thus each element of the merger coefficient vector represents the contribution of the corresponding column of the LoRA matrix $\Delta W$ to the final merge.

Our \name{} approach has two goals: \textbf{(1)} to minimize the interference between content and style LoRAs, defined by the cosine similarity between the columns of content and style LoRAs while \textbf{(2)} conserving the capability of the merged LoRA to generate the reference subject and style independently by minimizing the difference between subject/style images generated by the mixed LoRA and original subject/style LoRAs. To ensure that the columns that are merged with each other minimize signal interference, our proposed loss seeks to minimize the alignment between the merge vectors $m_c$ and $m_s$ of each layer. Meanwhile, we wish to ensure that the original behavior of both the style and the content LoRAs is preserved in the merged model. Therefore, as depicted in Fig.~\ref{fig:method_overview}, we formulate an optimization problem with following loss function:
\begin{align}
\mathcal{L}_{merge} =& \| (D \oplus L_m)(x_c, p_c) - (D \oplus L_c)(x_c, p_c)\|_2  \nonumber \\ 
    +& \| (D \oplus L_m)(x_s, p_s) - (D \oplus L_s)(x_s, p_s)\|_2 \nonumber \\
    + &\lambda \sum_{i} |m^{(i)}_c \cdot m^{(i)}_s|,
\label{eq:optfn}
\end{align}
where the merged model $L_m$ is calculated using $m_c$ and $m_s$ as per Eq.~\ref{eq:merge}; $(x_c, x_s)$ and $(p_c, p_s)$ are noisy latents and text conditioning prompts for content and style references respectively, and $\lambda$ is an appropriate multiplier for the cosine-similarity loss term. Note that the first two terms ensure that the merged model retains the ability to generate individual style and content, while the third term enforces an orthogonality constraint between the columns of the individual LoRA weights. Importantly, we keep the weights of the base model and the individual LoRAs frozen, and update only the merger coefficient vectors. As seen in the next section, such a simple optimization method is effective in producing strong stylization of custom subjects. Further, \name{} requires only $100$ gradient updates which is $10\times$ lower compared to joint training approaches.

\section{Experiments}
\label{sec:expt}
\begin{figure}[tp]
    \centering
    \includegraphics[width=0.99\linewidth]{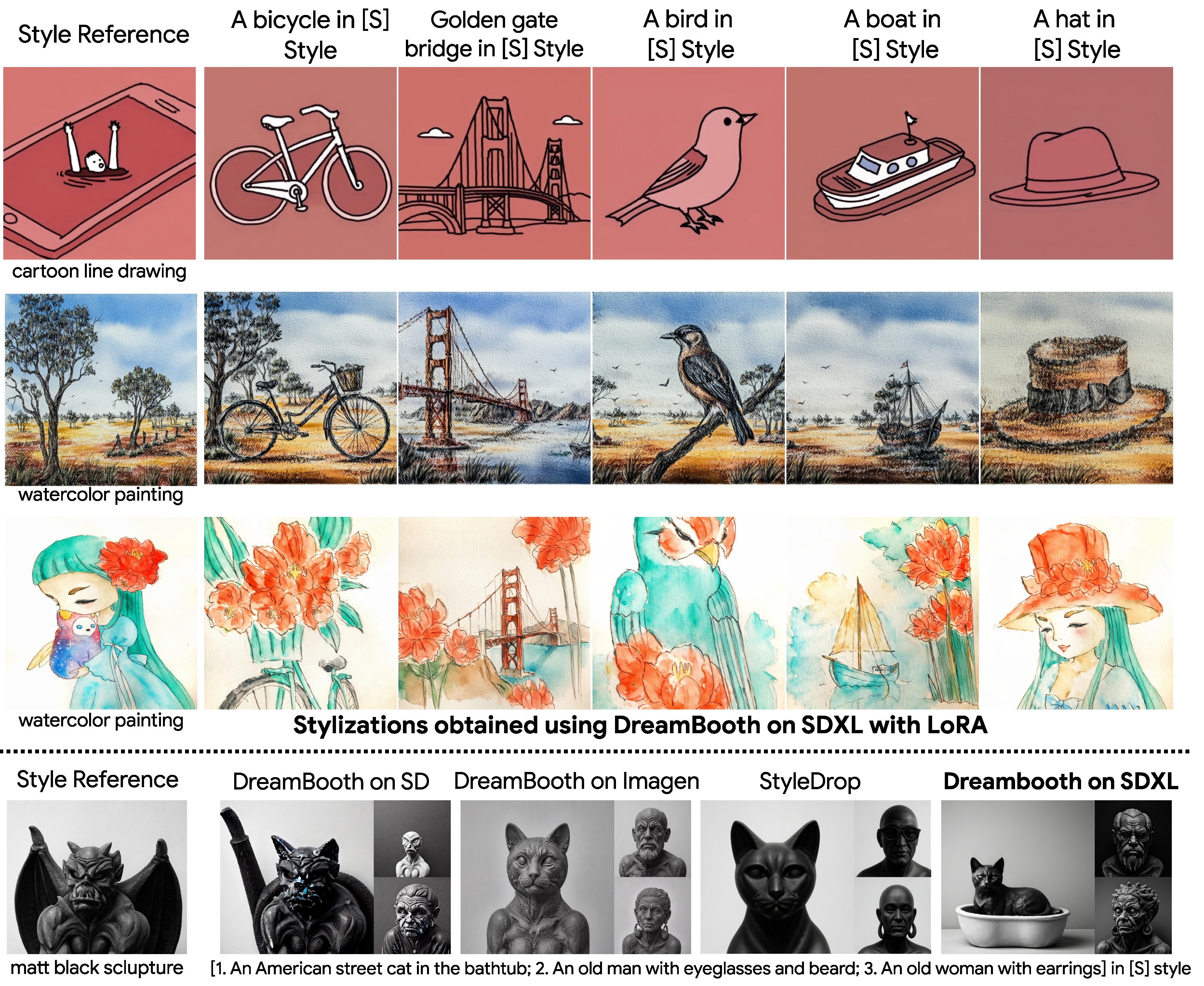}
    \caption{\textbf{Style Learning using DreamBooth on SDXL.} Top: SDXL model learns to produce stylized outputs when fine-tuned on a single example of a reference style using LoRA with a DreamBooth objective. Bottom: The stylizations produced by fine-tuned SDXL model are superior to those of other models.
    Note that unlike StyleDrop, SDXL DreamBooth fine-tuning does not require human feedback.}
    \label{fig:sdxl_styledrop}
    \vspace{-1em}
\end{figure}

\noindent \textbf{Datasets.} We choose a diverse set of content images from the DreamBooth dataset~\cite{ruiz2023dreambooth}, which provides 30 image sets each containing 4-5 images of a given subject. Similarly, a diverse set of style reference images is selected from the data provided by authors of StyleDrop~\cite{sohn2023styledrop}. We use only a single image for each style. The attribution and licence information for all the content and style images used are available in the DreamBooth and StyleDrop manuscripts/websites, and we also include them in the supplementary material. 

\noindent \textbf{Experimental Setup.} We perform all our experiments using the SDXL v1.0~\cite{Podell2023SDXLIL} base model. We use DreamBooth fine-tuning with LoRA of rank $64$ for obtaining all the style and content LoRAs. We update the LoRA weights using Adam optimizer for $1000$ steps with batch size of $1$ and learning rate of $0.00005$. We keep the text encoders of SDXL frozen during the LoRA fine-tuning. For ZipLoRA, we use $\lambda=0.01$ in Eq.~\ref{eq:optfn} for all our experiments, and run the optimization until cosine similarity drops to zero with a maximum number of gradient updates set to $100$. We plan to release the implementation of our method in future. To obtain qualitative and quantitative comparisons with existing methods, we use their official open-source implementations except for StyleDrop~\cite{sohn2023styledrop}. Since the official code and the model for StyleDrop is not available publicly, we obtain its results by contacting the authors.

\subsection{Style-tuning behavior of SDXL model}
\label{sec:styletune}
As discussed in Sec.~\ref{sec:method}, we observe, surprisingly, that a pre-trained SDXL model exhibits strong style learning when fine-tuned on only one reference style image. We show style-tuning results on SDXL model in Fig.~\ref{fig:sdxl_styledrop}. For each reference image, we apply LoRA fine-tuning of SDXL model using DreamBooth objective with LoRA rank$=64$. For fine-tuning, we follow a similar prompt formation as provided in StyleDrop: ``an $<$object$>$ in the $<$style description$>$ style". Once fine-tuned, SDXL is able to represent diverse set of concepts in the reference style by capturing the nuances of painting style, lighting, colors, and geometry accurately. The question of why this model exhibits this strong style learning performance, as opposed to the lesser performance of previous Stable Diffusion versions~\cite{Rombach2021HighResolutionIS} (or Imagen~\cite{Saharia2022PhotorealisticTD}) is left open and can have many answers including training data, model architecture and training schemes.

We also provide comparisons with StyleDrop on Muse~\cite{chang2023muse}, DreamBooth on Imagen, and DreamBooth on Stable Diffusion (SDv1.5) in Fig.~\ref{fig:sdxl_styledrop}. We observe that SDXL style-tuning performs significantly better than the competing methods. Note that StyleDrop requires iterative training with human feedback whereas SDXL style-tuning does not. 
This behavior of SDXL makes it the perfect candidate for investigating the merging of style LoRAs with subject LoRAs to achieve personalized stylizations. Thus, we choose to use it as a base model for all of our experiments.

\subsection{Personalized Stylizations}
\begin{figure*}[tp]
    \centering
    \includegraphics[width=0.99\linewidth]{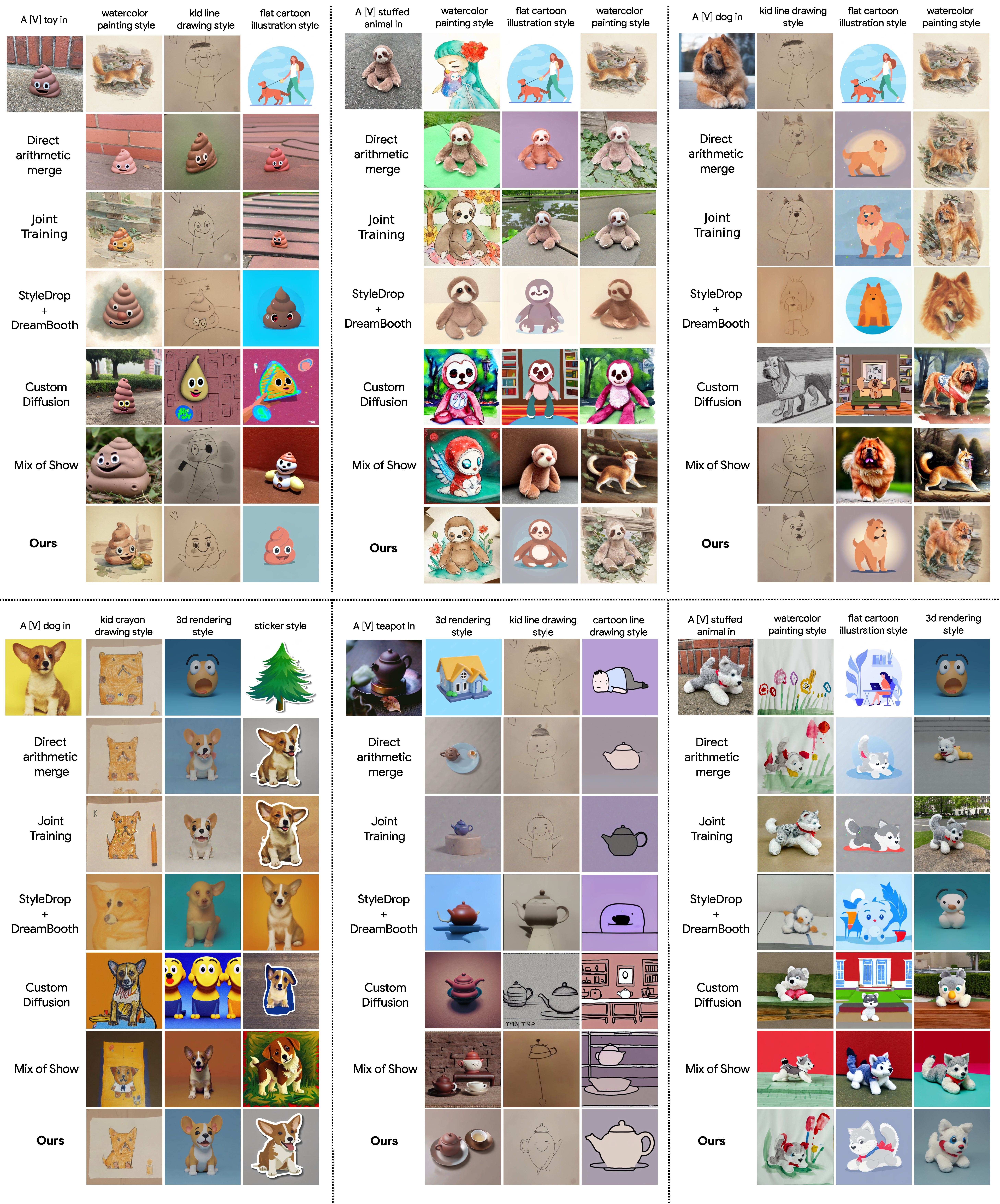}
    \caption{
    \textbf{Qualitative Comparison:} We compare samples from our method (Ours), versus direct arithmetic merge~\cite{simo_lora}, joint training, StyleDrop with DreamBooth~\cite{sohn2023styledrop}, Mix of Show~\cite{gu2023mixofshow}, and Custom Diffusion~\cite{kumari2022customdiffusion}. We observe that our method achieves strong style and subject fidelity that surpasses competing methods. We provide additional results in Supplementary.
    }
    \label{fig:compare}
\end{figure*}

To start with, we obtain the style LoRAs following the style-tuning on SDXL as described in Sec.~\ref{sec:styletune}, and obtain object LoRAs by applying DreamBooth fine-tuning on the subject references. Fig.~\ref{fig:teaser} and Fig.~\ref{fig:compare} show the results of our approach for combining various style and content LoRAs. Our method succeeds at both preserving the identity of the reference subject and capturing the unique characteristics of the reference style.

We also present qualitative comparisons with other approaches in Fig.~\ref{fig:compare}. As a baseline, we compare with the direct arithmetic merge~\cite{simo_lora} obtained through Eq.~\ref{eq:direct} with $w_c$ and $w_s$ set to 1. Such direct addition results in loss of information captured in each LoRA and produces inferior results with distorted object and/or style. 

We additionally compare our method with joint training of subject and style using a multi-subject variant of DreamBooth with multiple rare unique identifiers. As shown, joint training fails to learn the disentanglement between object and style and produces poor results. It also is the least flexible method since it does not allow the use of pre-trained LoRAs, neither can it be used as a style-only or content-only LoRA. Further, it requires $10\times$ as many training steps as ZipLoRA.

\begin{figure*}
	\begin{minipage}{0.45\linewidth}
		\begin{center}
			\centering
			\setlength{\tabcolsep}{4pt}
			\captionof{table}{\small\textbf{User Preference Study}. We compare the user preference of accurate stylization and subject fidelity between our approach and competing methods. Users generally prefer our approach.}
			\label{tab:userstudy}
			\vspace{-0.5em}
			{\scalebox{0.78}{\begin{tabular}{lccccc}
						\toprule
						\multicolumn{6}{c}{\textbf{$\%$ Preference for \name{} over:}} \\
						\cmidrule(lr){1-6}
						& Direct Merge~~& \makecell{Joint \\Training} & \makecell{StyleDrop+\\DreamBooth} & \makecell{Mix of Show} & \makecell{Custom \\Diffusion} \\
						\midrule
						&$82.7\%$&$71.1\%$&$68.0\%$ & $87.3\%$ & $88.1\%$ \\
						\bottomrule
			\end{tabular}}}
		\end{center}
	\end{minipage}
	\quad
	\begin{minipage}{0.52\linewidth}
		\centering
		\captionof{table}{\small\textbf{Alignment Scores.} We compare cosine similarities between CLIP (for style \& text) and DINO (for subject) features of the output and reference style, subject, and prompt respectively. ZipLoRA achieves superior subject \& text fidelity while also maintaining the style alignment.}
		\label{tab:alignscore}
		\vspace{-0.5em}
		{\scalebox{0.7}{\begin{tabular}{lcccccc}
					\toprule
					& ZipLoRA & \makecell{Joint\\ Training} & \makecell{Direct\\ Merge} & \makecell{StyleDrop +\\DreamBooth} & \makecell{Mix of \\Show} & \makecell{Custom \\ Diffusion} \\
					\midrule
					Style-alignment & $0.699$ & $0.680$ & $0.702$ & $0.646$ & $0.635$ & $0.616$\\
					Subject-alignment & $0.420$ & $0.378$ & $0.357$ & $0.394$& $0.374$  & $0.346$\\
					Text-alignment & $0.303$ & $0.296$ & $0.275$ & $0.263$ & $0.251$ &$0.262$\\
					\bottomrule
		\end{tabular}}}
	\end{minipage}
\end{figure*}

StyleDrop~\cite{sohn2023styledrop} proposes a StyleDrop+DreamBooth approach for achieving personalized stylizations, where a StyleDrop method is applied on a DreamBooth model fine-tuned on the reference object. Our comparisons show that its performance is not ideal, considering the high compute cost and human feedback requirements. It also requires adjusting the object and style model weights $w_c$ and $w_s$ similar to the direct merge in order to produce reasonable outputs, while our method is free from any such hyperparameter tuning.

Further, we compare our method with recent multi-concept generation approaches Mix of Show~\cite{gu2023mixofshow} and Custom Diffusion~\cite{kumari2022customdiffusion}. Our results reveal that both the methods perform inferior to \name{}. Mix of show relies on region-aware prompting that requires spatial disentanglement between the individual concepts, thus performs poorly for subject-style merge since the style is usually spread across the entire image. Moreover, it uses a custom LoRA model referred as ED-LoRA thus requires training from scratch for each individual concept. Custom Diffusion learns unique text tokens for each concept which does not work reliably when it comes to combining a style with a subject.

\noindent\textbf{Quantitative results. }We conduct user studies for a quantitative comparison of our method with existing approaches. In our study, each participant is shown a reference subject and a reference style along with outputs of two methods being compared, in a random order, and asked which output best depicts the reference style while preserving the reference subject fidelity. We conducted separate user studies for ZipLoRA vs. each of the five competing approaches, and received $360$ responses across $45$ users for each case. We show the results in Tab.~\ref{tab:userstudy}. As we can see, ZipLoRA receives higher user preference in all three cases owing to its high-quality stylization while preserving subject integrity.

\noindent Following DreamBooth~\cite{ruiz2023dreambooth}, we also provide comparisons using image-alignment and text-alignment scores in Tab.~\ref{tab:alignscore}. We employ three metrics: for style-alignment, we use CLIP-I scores of image embeddings of output and the style reference; for subject-alignment, we employ DINO features for the output and the reference subject; and for text-alignment, we use CLIP-T embeddings of the output and the text prompt. In all three cases, we use cosine-similarity as the metric and calculate averages over 4 subjects in 8 styles each. ZipLoRA results in competitive style-alignment scores as compared to joint training and direct merge, while achieving significantly better scores for subject-alignment. This highlights ZipLoRA's superiority in maintaining the subject fidelity. ZipLoRA also outperforms the existing methods in text-alignment, implying that it preserves the text-to-image generation capability, and also expresses the designated style and subject better (since these are also part of the text prompt). One should note that these metrics are not perfect, particularly when it comes to measuring style alignment, since they lack the ability to capture subtle stylistic details, and are entangled with semantic properties of images, such as the overall content.
\begin{figure*}[!h]
	\centering
	\includegraphics[width=1\linewidth]{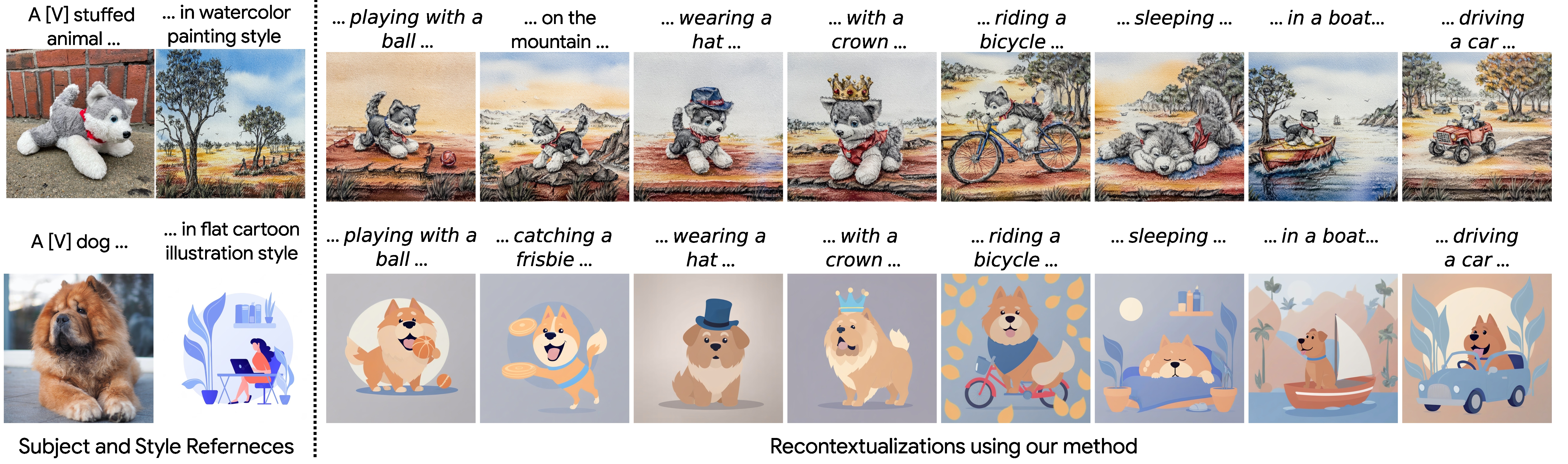}
	\caption{Our method successfully re-contextualizes the reference subject while preserving the stylization in the given style.}
	\label{fig:recontext}
\end{figure*}

\begin{figure}[tp]
\centering
\begin{subfigure}{.45\textwidth}
  \centering
  \includegraphics[width=0.99\linewidth]{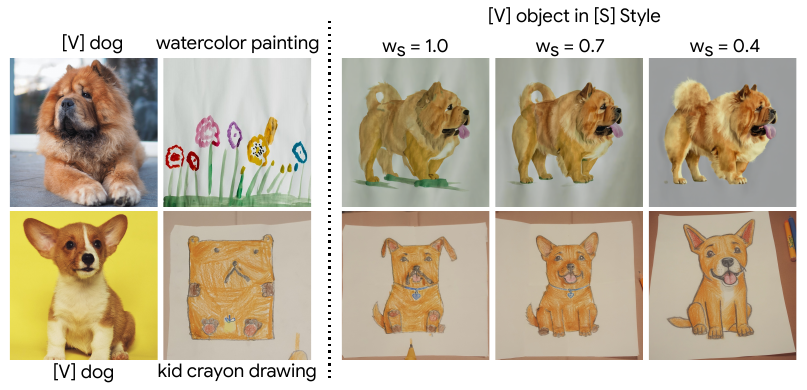}
    \caption{\scriptsize\textbf{Style Controllability of ZipLoRA}}
    \label{fig:style_control}
\end{subfigure}%
\quad
\begin{subfigure}{.45\textwidth}
  \centering
  \includegraphics[width=0.99\linewidth]{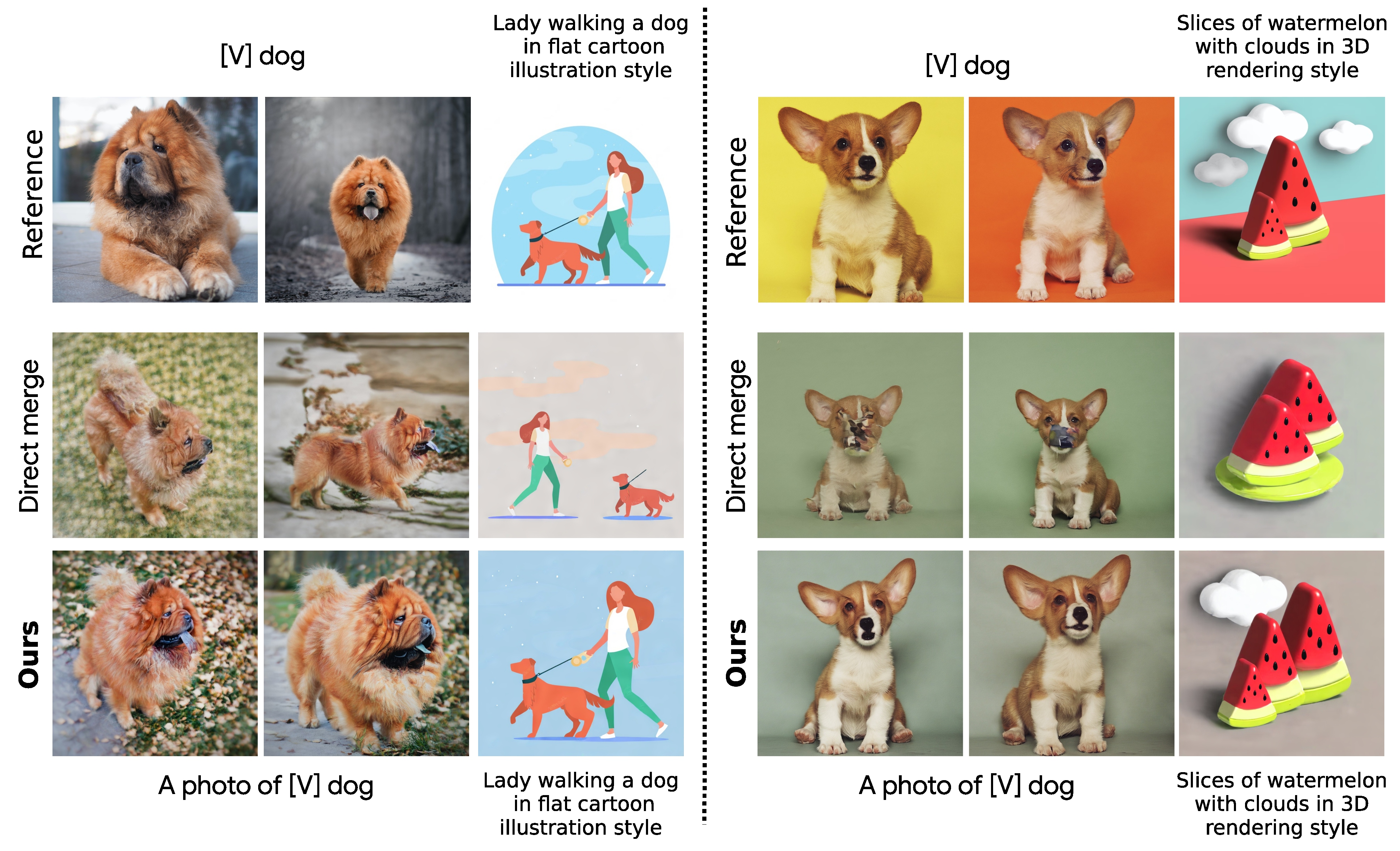}
    \caption{\scriptsize \textbf{Retention of the individual concepts}}
    \label{fig:moe}
\end{subfigure}
\caption{\textbf{(a) }Our method works out-of-the-box at achieving good subject and style personalization. Nevertheless, varying the merging weights $w_s$ allows for controlling the extent of stylization. \textbf{(b)} Our method does not lose the ability to generate individual concepts, unlike the direct merge approach.}
\end{figure}

\noindent \textbf{Ability to re-contextualize. }The merged \name{} model can recontextualize reference objects in diverse contexts and with semantic modifications while maintaining stylization quality. As shown in Fig.~\ref{fig:recontext}, our method preserves the base model's text-to-image generation capabilities while accurately stylizing the entire image in the reference style. Such ability is highly valuable in various artistic use cases that requires controlling contexts, subject identities, and styles. 

\noindent \textbf{Controlling the extent of stylization. }Our optimization-based method directly provides a scalar weight value for each column of the LoRA update, thus eliminating a need for tuning and adjustments for obtaining reasonable results. However, we can still allow the strength of object and style content to be varied for added controllability. One can attenuate the style layer weights by multiplying them with an additional scalar multiplier $w_s$ to limit the contribution of the style in the final output. As shown in Fig.~\ref{fig:style_control}, this allows for a smooth control over the extent of stylization as $w_s$ varies between $0$ to $1$.

\noindent \textbf{Ability to produce the reference object and the style.} Apart from producing accurate stylizations, an ideal LoRA merge should also preserve the ability to generate individual object and style correctly. This way, a merged LoRA model can also be used as a replacement of both the individual LoRAs. As shown in Fig.~\ref{fig:moe}, our approach retains the original behavior of both the models and can accurately generate specific structural and stylistic elements of each constituent LoRA, while direct merge fails.

\begin{figure}[tp]
	\begin{center}
		\includegraphics[width=1.0\linewidth]{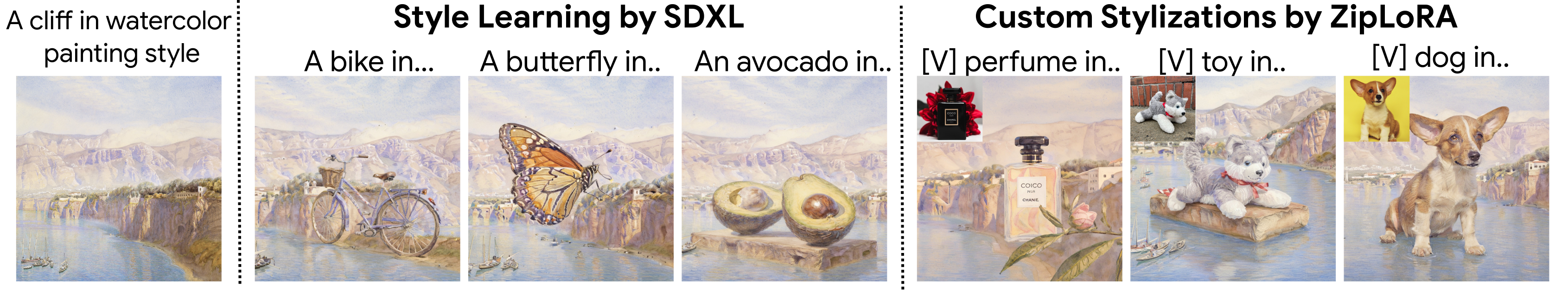}
	\end{center}
    \vspace{-1.0em}
	\caption{\small{ \textbf{Failure Cases. }For a few styles, \name{} fails to separate the content of the style reference from its style, resulting into the leakage of the content (cliff in this case) in stylization outputs.}}
	\label{fig:fail}
 \vspace{-1.8em}
\end{figure}

\noindent\textbf{Limitations/failure cases.} For some style reference images, instead of capturing just the style, SDXL style-tuning incorrectly captures the subject as well. \name{} fails to disentangle such styles further, thus the content of style reference can leak into the stylization outputs. For example, as shown in Fig.~\ref{fig:fail}, SDXL style-tuning fails to disentangle the cliff from the watercolor painting style, and \name{} ends up producing the cliff in the background in all the stylizations.

\noindent\textbf{Comparisons of runtime/storage.} \name{} offers improved efficiency, exhibiting lower storage footprints, reduced computational demands, and faster runtimes. \name{} requires only $100$ gradient updates which is $10\times$ less than Joint Training (JT), Custom Diffusion (CD), and Mix of Show (MoS). \name{}'s runtime is $560$ seconds while JT and CD takes $3540$s and $3890$s respectively. For MoS, to achieve a successful merger, one first needs to obtain ED-LoRAs for each individual concept, thus the total runtime for MoS is $4980$s ($1600$s each for training ED-LoRAs + $1780$s for merging them). All runtimes are calculated on single NVIDIA A100. \name{} updates only the merger coefficient vectors $m_c, m_s$ while keeping the LoRA weights frozen, thus has only $1.6$M trainable parameters as opposed to $180$M in the case of the competing methods, reducing the GPU memory requirements from $38$GB to $21$GB. For storage, \name{} needs to store just the merger coefficient vectors $m_c, m_s$ requiring only $6.5$MB of storage, while the LoRA resulting from other methods requires $360$MB.

\noindent\textbf{Performance on Stable Diffusion (SDv1.5).} As discussed in Sec.~\ref{sec:styletune} \& Fig.~\ref{fig:sdxl_styledrop}, LoRA fine-tuning on earlier version of Stable Diffusion (SDv1.5) fails to capture the stylization, thus the performance of \name{} on SDv1.5 becomes limited by the stylization ability of the underlying style LoRA. Our successful style-tuning of SDXL is key observation that led us to adopt it as the base model. That being said, our observations about sparsity and alignment of LoRA weights remain valid for other models of stable diffusion family, and even on SDv1.5, ZipLoRA outperforms competing methods (Direct Merge, Joint Training, Custom Diffusion, and Mix of Show) by achieving better stylizations with improved subject and style fidelity. We provide comparison figure and quantitative results on SDv1.5 in Supplementary.
\vspace{-0.5em}
\section{Conclusion and Future Work}
\vspace{-0.5em}
\label{sec:conclusion}
In this work, we have introduced \textbf{\name{}}, a novel method for seamlessly merging independently trained style and subject LoRAs. Our approach unlocks the ability to generate \textbf{any subject in any style} using contemporary diffusion models like SDXL. By leveraging key insights about pre-trained LoRA weights, we surpass existing methods for this task. \name{} offers a streamlined, cheap, and hyperparameter-free solution for simultaneous subject and style personalization, unlocking a new level of creative controllability for diffusion models. While \name{} focuses on merging a pair of a subject and a style LoRA, combining more than two subject/style LoRAs can be considered as a future work.

\noindent\textbf{Acknowledgements.} We thank Prafull Sharma, Meera Hahn, Jason Baldridge, and Dilip Krishnan for helpful discussions and suggestions. We also thank Kihyuk Sohn for helping with the generation of StyleDrop results.
{
    \small
    \bibliographystyle{ieeenat_fullname}
    \bibliography{main,vsbib}
}

 \newpage
\section*{Appendix}
\renewcommand{\thesection}{\Alph{section}}
\renewcommand{\thesubsection}{\Alph{section}.\arabic{subsection}}
\setcounter{section}{0}
\section{Additional Implementation Details}
\label{sec:impl}
In this section, we provide additional implementation details for our algorithm:
\begin{itemize}
    \item We train both the base LoRAs corresponding to style and content using standard DreamBooth protocol on SDXL made available by diffusers python library~\cite{von-platen-etal-2022-diffusers}. For training, we use $1000$ fine-tuning steps with batch size $1$ and learning rate $5e-5$. We do not train text encoders during such fine-tuning. Further, we use rank$=64$ for obtaining both the style and content LoRAs.
    \item We do not use SDXL refiner model in any of our experiments, neither for training nor for inference. 
    \item For ZipLoRA, we initialize the merger coefficient vectors with all ones. This is a natural way to initialize, since it imitates the direct merge at the beginning of the training, and gradually update the merger coefficients to minimize the alignment term along with maintaining the capability to generate both the individual concepts.
    \item For \name{} fine-tuning, we use $\lambda=0.01$ and learning rate $0.01$ for all our experiments.
    \item We keep the number of diffusion inference steps fixed to $50$ in all our experiments.
\end{itemize}

\section{Performance of \name{} on Stable Diffusion}
As discussed in the main paper, LoRA fine-tuning on earlier version of Stable Diffusion (SDv1.5) fails to capture the stylization faithfully, thus the performance of \name{} on SDv1.5 becomes limited by the stylization ability of the underlying style LoRA. That being said, our observations about sparsity and alignment of LoRA weights remain valid for other models of stable diffusion family, and even on SDv1.5, ZipLoRA outperforms competing methods (Direct Merge, Joint Training, Custom Diffusion, and Mix of Show) by achieving better stylizations with improved subject and style fidelity. 

\begin{figure*}[tp]
    \centering
    \includegraphics[width=0.7\linewidth]{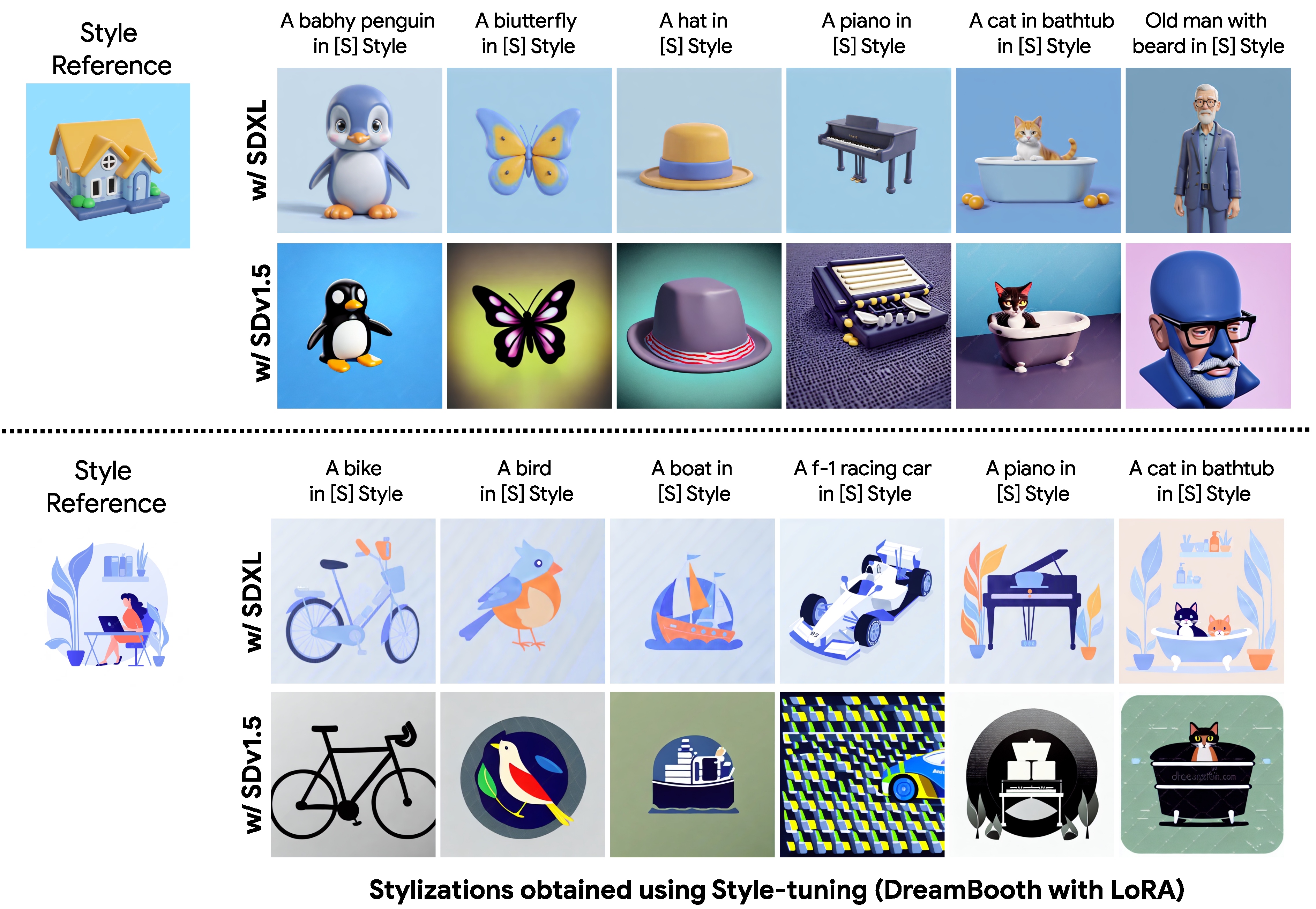}
    \caption{\textbf{Comparison between SDv1.5 and SDXL for Style Learning using DreamBooth with LoRA.} SDXL model (top row) produces superior quality outputs when fine-tuned on a single example of a reference style (left-most column) using LoRA with a DreamBooth objective. Notice that SDv1.5 (bottom row) fails to capture the reference style consistently.}
    \label{fig:style_tuning_s}
\end{figure*}

In this regard, we provide additional style-tuning results on SDv1.5 model in Fig.~\ref{fig:style_tuning_s}. One can see that the quality of the stylizations captured by SDv1.5 model is underwhelming as compared to that of SDXL. 

\begin{figure*}[h]
	\begin{center}
		\includegraphics[width=0.70\linewidth]{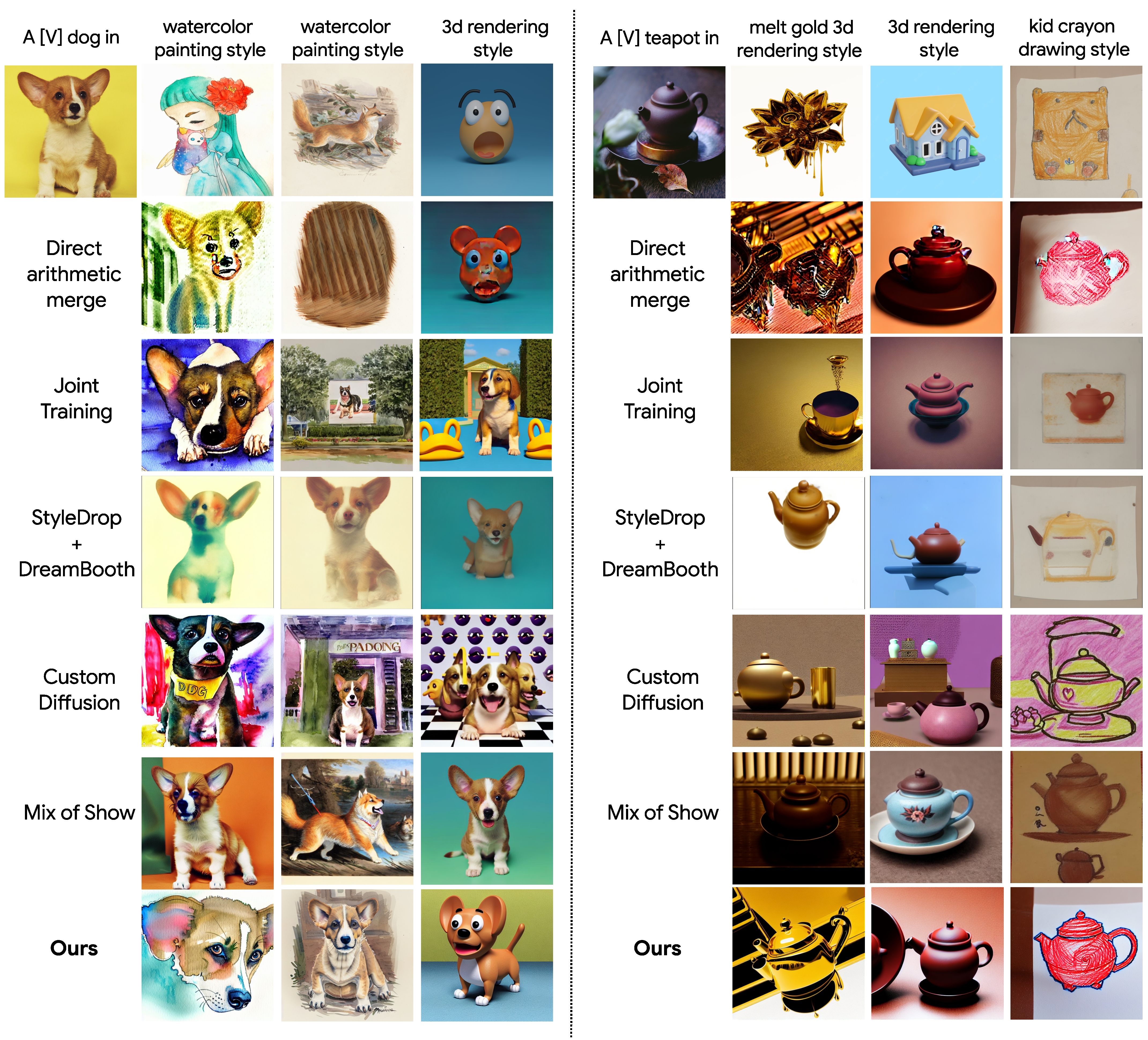}
	\end{center}
	\caption{\textbf{Performance of \name{} on SDv1.5. }Even on SDv1.5, \name{} outperforms Direct Merge, Joint Training, StyleDrop+DreamBooth, Custom Diffusion, and Mix of Show.}
	\label{fig:compare_sd}
\end{figure*}

Further, we also obtain custom stylizations by merging subject and style LoRAs using \name{} on SDv1.5, and compare the results with Direct Merge, Joint Training, Custom Diffusion, and Mix of Show in Fig.~\ref{fig:compare_sd}. Note that we use SDv1.5 as a base model for these competing methods as well. For completeness, we also include the results for StyleDrop+DreamBooth. Note that StyleDrop is model-specific method that uses Muse as the base model, and since its code is not public, it is not possible to evaluate it on SDv1.5. As one can see in Fig.~\ref{fig:compare_sd}, even on SDv1.5, \name{} produces superior stylization outputs and surpass all the competing methods.

We also provide quantitative evaluations on subject, style, and text alignment for our method and competing methods for SDv1.5 in Tab.~\ref{tab:sd_s}.

\begin{figure}[b]
	\centering
	\includegraphics[width=0.5\linewidth]{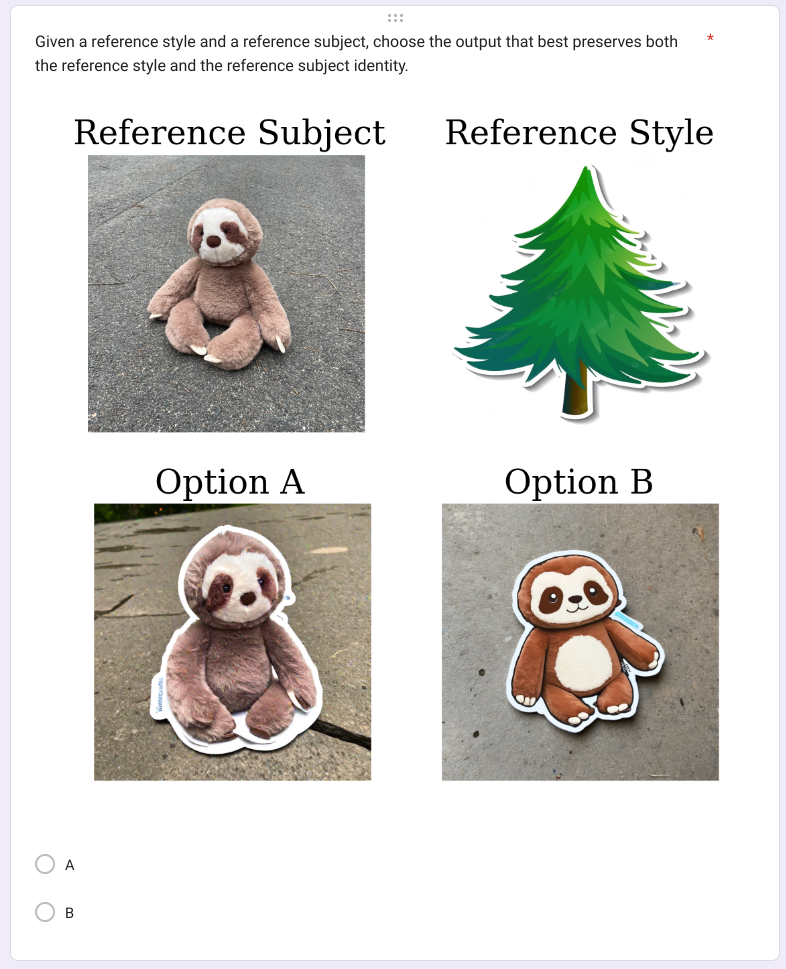}
	\caption{\textbf{User Interface of Our User Studies.} Each participant is shown a reference subject and a reference style along with outputs of two methods being compared and asked which output best depicts the reference style while preserving the reference subject fidelity.}
	\label{fig:userstudy}
\end{figure}

\section{Additional Results}
\label{sec:expt_results}
We provide additional results for experiments discussed in the main paper to present supporting evidence for the claims made.
\subsection{Qualitative comparisons for personalized stylization on SDXL. }
\begin{figure*}[tp]
    \centering
    \includegraphics[width=0.75\linewidth]{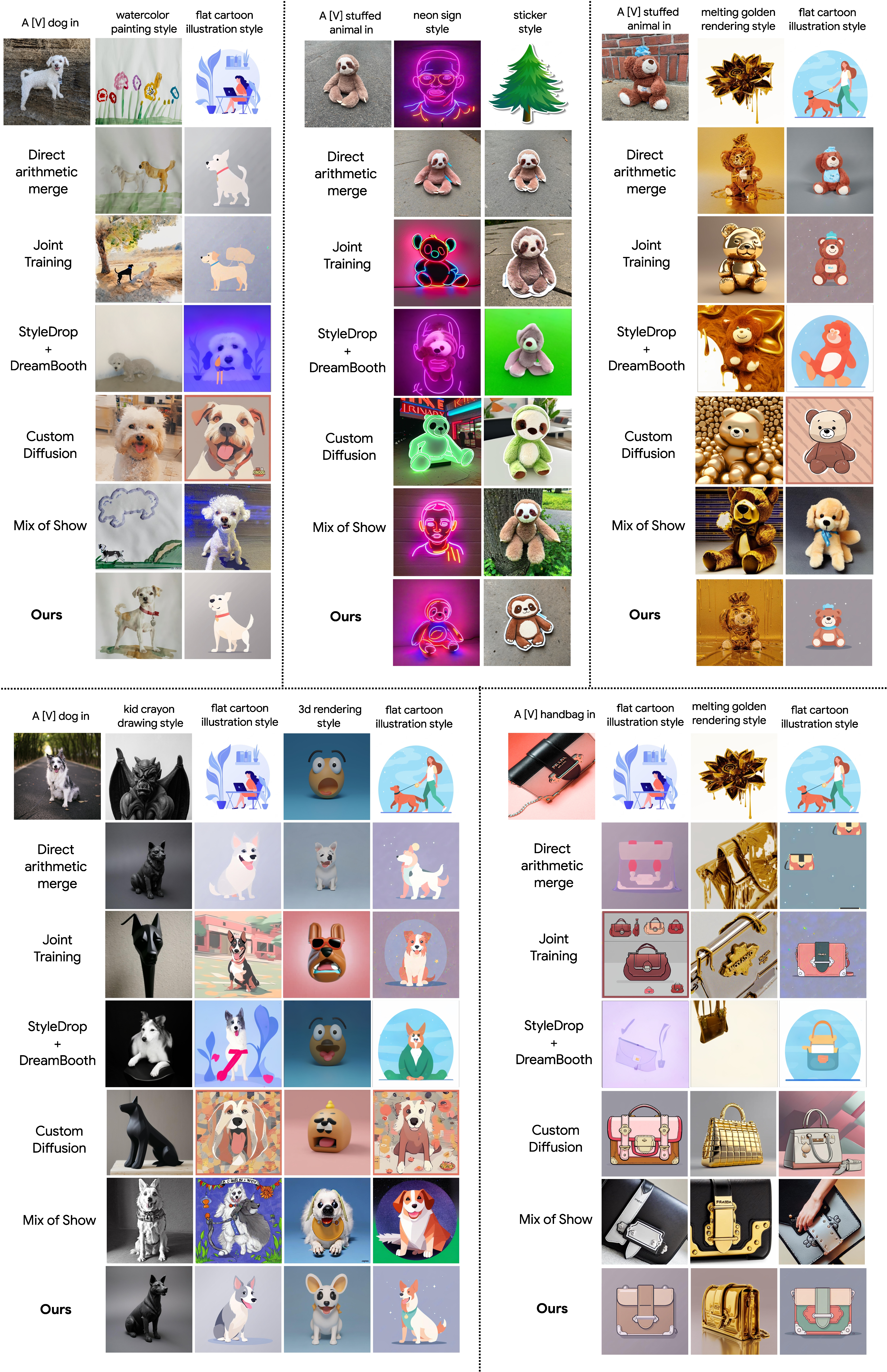}
    \caption{
    \textbf{Additional Qualitative Comparisons for Personalized Stylizations using \name{} on SDXL.} We compare samples from our method (Ours), versus Direct Arithmetic Merge, Joint Training, StyleDrop+DreamBooth, Custom Diffusion, and Mix of Show. We observe that our method achieves strong style and subject fidelity that surpasses competing methods.
    }
    \label{fig:compare_s}
\end{figure*}

We provide additional qualitative comparison of our method with Direct Merge, Joint Training, StyleDrop+DreamBooth, Custom Diffusion, and Mix of Show in Fig.~\ref{fig:compare_s}. Superior results obtained using ZipLoRA further strengthens the claims of improved performance over the baselines. 
\subsection{Evidence that LoRA updates are sparse.} 
\begin{figure*}[tp]
    \centering
    \includegraphics[width=0.75\linewidth, trim={2pt 0pt 0pt  0pt}, clip]{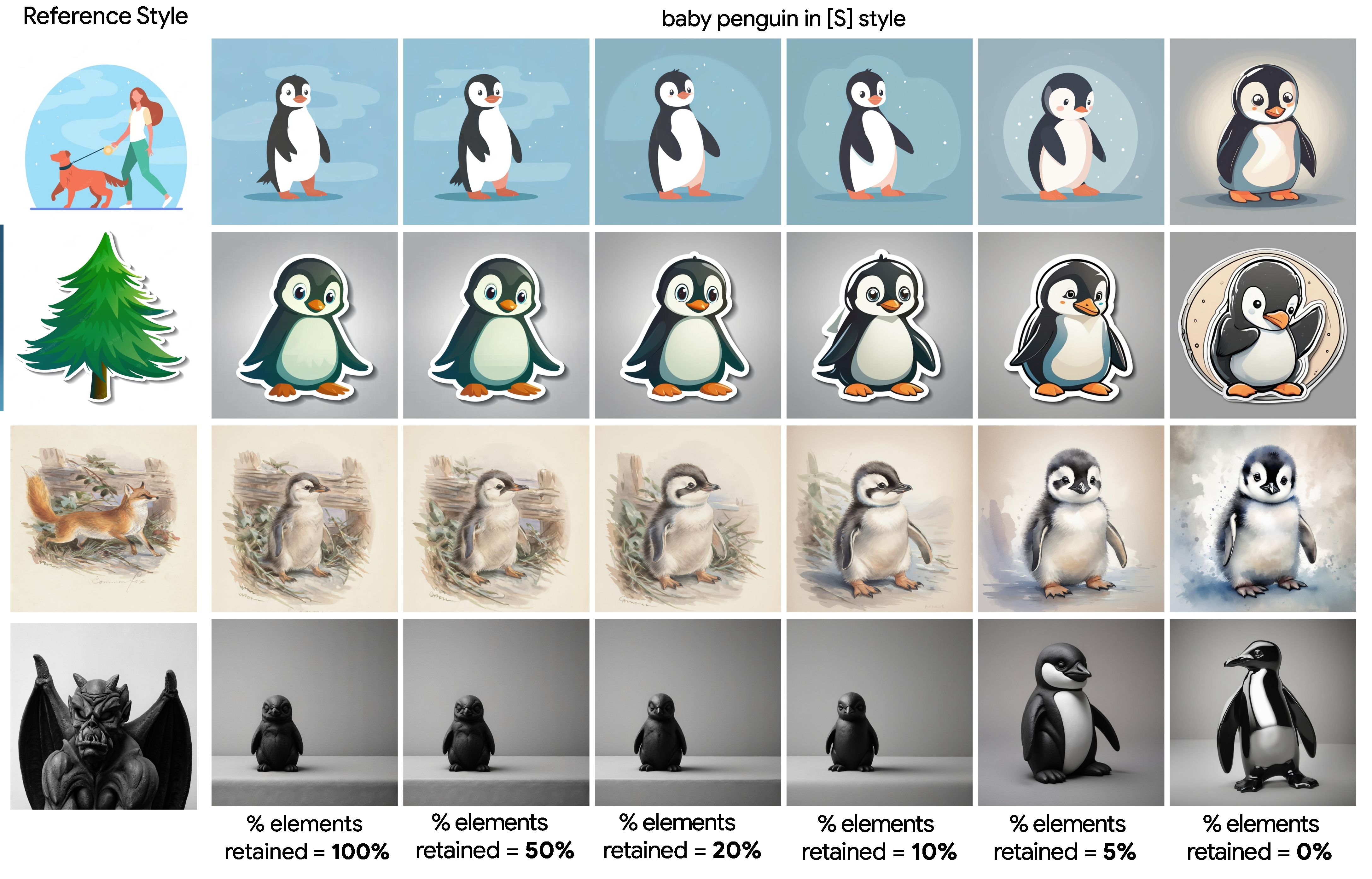}
    \caption{
    \textbf{Additional Results: LoRA weight matrices are sparse.} Most of the elements in $\Delta W$ have a magnitude very close to zero, and can be conveniently thrown away without affecting the generation quality of the fine-tuned model. The stylization quality is maintained even when only $20\%$ of the elements are retained.
    }
    \label{fig:sparse_s}
\end{figure*}

In Fig.~\ref{fig:sparse_s}, we present more evidence for our claim that the LoRA updates are sparse in general, and significant chunk of low magnitude elements can be thrown away without affecting the stylization performance. As one can see, the stylization performance remains unaffected even when $80\%$ of the elements are thrown away, while stylization degrades if this number is increased further. 

\subsection{Additional results of style-tuning using SDXL. }
\begin{figure*}[bp]
    \centering
    \includegraphics[width=0.75\linewidth]{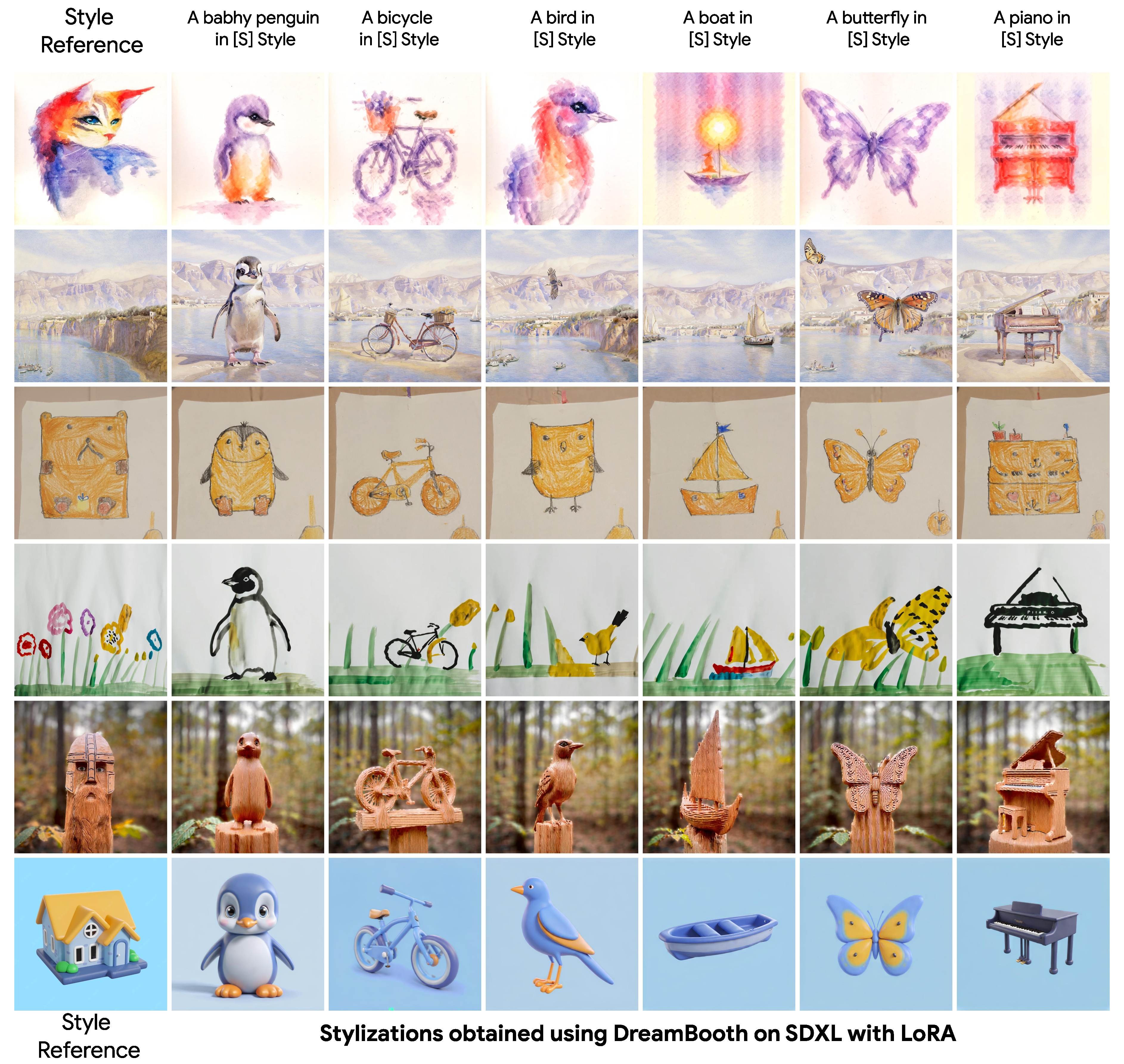}
    \caption{\textbf{Additional Results for Style Learning using DreamBooth on SDXL.} SDXL model learns to produce stylized outputs when fine-tuned on a single example of a reference style (left-most column) using LoRA with a DreamBooth objective. 
    Note that unlike StyleDrop, SDXL DreamBooth fine-tuning does not require human feedback.}
    \label{fig:sdxl_styledrop_s}
\end{figure*}

We also provide additional results on Style-tuning property of SDXL model in Fig.~\ref{fig:sdxl_styledrop_s}. As shown, SDXL model can learn to generate stylized images in a given style through simple application of DreamBooth method without requiring any human feedback.

\begin{table*}[bp]
	\begin{center}
				\centering
				\setlength{\tabcolsep}{4pt}
				\captionof{table}{\textbf{Alignment Scores for \name{} on SDv1.5.} While the stylization capabilities of SDv1.5 are inferior to SDXL, ZipLoRA still provides superior subject and text fidelity as compared to the existing methods when used on SDv1.5.}
				\label{tab:sd_s}
				\scalebox{0.8}{\begin{tabular}{lcccccc}
						\toprule
						& \makecell{ZipLoRA \\ \\(on SDv1.5)} & \makecell{Joint\\ Training \\ (on SDv1.5)} & \makecell{Direct\\ Merge \\ (on SDv1.5)} & \makecell{Mix of \\Show \\(on SDv1.5)} & \makecell{Custom \\ Diffusion \\ (on SDv1.5)} &\makecell{StyleDrop+\\DreamBooth \\ (on Muse)}\\
						\midrule
						Style-alignment $\uparrow$ & $\mathbf{0.651}$ & $0.579$  & $0.581$ & $0.618$  & $0.574$ & $0.646$\\
						Subject-alignment $\uparrow$ & $\mathbf{0.413}$ & $0.235$ & $0.222$& $0.323$ & $0.311$ & $0.394$\\
						Text-alignment $\uparrow$ & $\mathbf{0.283}$ & $0.247$ &  $0.241$&  $0.221$ &$0.237$ & $0.263$\\
						\bottomrule
				\end{tabular}}
			\end{center}
		\end{table*}
	
\section{User Studies Details}
We conduct user studies for a quantitative comparison of our method with existing approaches. We cast it as a binary comparison task thus conduct separate study for each pair of methods. We used Google Forms to conduct the user studies (See the user interface of our study in Fig.~\ref{fig:userstudy}). In our study, each participant is shown a reference subject and a reference style along with outputs of two methods being compared and asked which output best depicts the reference style while preserving the reference subject fidelity. Options A and B are flipped randomly for each question. Participants for the study are selected at random from a pool of volunteers. For every study, each participant is asked $8$ questions, thus each participant answers $40$ questions in total across $5$ studies that we conducted. $45$ participants responded to our study, resulting in $360$ responses for each of the five studies ($1800$ responses in total). As indicated in the results table in the main paper, our method is preferred over completing methods in all the five studies.

\section{Societal Impact} This project empowers users to personalize both the subject and the artistic style of their images, featuring individual subjects like animals or objects, and styles like watercolor or sketch. It is important to acknowledge that, similar to other generative models and image editing techniques, this technology could be misused to create deceptive content. Addressing these potential ethical concerns remains an ongoing priority in the field of generative modeling, particularly with regards to image manipulation.
\section{Datasets and Image Attributions}
\label{sec:attr}
We use style and content images from the datasets collected by StyleDrop~\cite{sohn2023styledrop} and DreamBooth~\cite{ruiz2023dreambooth} respectively. Note that these datasets do not contain any human subjects data or personally identifiable information. We provide image attributions below for each image that we used in our experiments. We refer readers to manuscripts and project websites of StyleDrop and DreamBooth for more detailed information about the usage policy and licensing of these images. 

\subsection{Image attributions for style references}
StyleDrop project webpage provides the image attribution information here:  \href{https://github.com/styledrop/styledrop.github.io/blob/main/images/assets/data.md}{Style Image Attribution} 

Specifically, the sources of the style images that we used in our experiments are as follows (linked as hyperlinks):

\href{https://unsplash.com/photos/0e6nHU8GRUY}{1}, 
\href{https://unsplash.com/photos/t0Bv0OBQuTg}{2}, 
\href{https://unsplash.com/photos/pink-yellow-and-green-flower-decors-6dY9cFY-qTo}{3}, 
\href{https://www.freepik.com/free-psd/three-dimensional-real-estate-icon-mock-up_32453229.htm}{4}, 
\href{https://it.freepik.com/vettori-gratuito/adesivo-albero-di-pino-su-sfondo-bianco_20710341.htm}{5}, 
\href{https://www.freepik.com/free-vector/young-woman-walking-dog-leash-girl-leading-pet-park-flat-illustration_11236131.htm}{6}, 
\href{https://unsplash.com/photos/t0Bv0OBQuTg}{7}, 
\href{https://unsplash.com/photos/0pJPixfGfVo}{8}, 
\href{https://unsplash.com/photos/H9g_HE6ZgGA}{9}, 
\href{https://unsplash.com/photos/jI3Lp0FYEz0}{10}, 
\href{https://unsplash.com/photos/kHuCUkkExbc}{11}, 
\href{https://www.instagram.com/p/Ck1tX2kvQrK/}{12}, 
\href{https://www.instagram.com/p/CqwU1bavm0T/}{13}, 
\href{https://img.freepik.com/free-vector/biophilic-design-workspace-abstract-concept_335657-3081.jpg}{14}, 
\href{https://unsplash.com/photos/a-golden-flower-with-drops-of-liquid-on-it-Prx96KdmWj0}{15}, 
\href{https://github.com/styledrop/styledrop.github.io/blob/main/images/assets/image_6487327_crayon_02.jpg}{16}, 
\href{https://unsplash.com/photos/gargoyle-statue-gZzUo--BTZ4}{17},
\href{https://unsplash.com/photos/a-wooden-carving-of-a-man-with-a-beard-CuWq_99U0xs}{18}

\subsection{Image attributions for content references}
DreamBooth project webpage provides the image attribution information here:
\href{https://github.com/google/dreambooth/blob/main/dataset/references_and_licenses.txt}{Content Image Attribution}

Specifically, the sources of the content images that we used in our experiments are as follows (linked as hyperlinks):
 
\href{https://github.com/google/dreambooth/tree/main/dataset/dog3}{1}, 
\href{https://github.com/google/dreambooth/tree/main/dataset/dog6}{2}, 
\href{https://github.com/google/dreambooth/tree/main/dataset/duck_toy}{3}, 
\href{https://github.com/google/dreambooth/tree/main/dataset/bear_plushie}{4}, 
\href{https://github.com/google/dreambooth/tree/main/dataset/grey_sloth_plushie}{5}, 
\href{https://github.com/google/dreambooth/tree/main/dataset/poop_emoji}{6}, 
\href{https://github.com/google/dreambooth/tree/main/dataset/dog5}{7}, 
\href{https://github.com/google/dreambooth/tree/main/dataset/robot_toy}{8}, 
\href{https://github.com/google/dreambooth/tree/main/dataset/teapot}{9}, 
\href{https://github.com/google/dreambooth/tree/main/dataset/vase}{10}, 
\href{https://github.com/google/dreambooth/tree/main/dataset/dog7}{11}, 
\href{https://github.com/google/dreambooth/tree/main/dataset/dog8}{12}, 
\href{https://github.com/google/dreambooth/tree/main/dataset/wolf_plushie}{13}, 
\href{https://github.com/google/dreambooth/tree/main/dataset/can}{14}, \href{https://github.com/google/dreambooth/tree/main/dataset/dog2}{15}

\end{document}